\pdfoutput=1

\documentclass[11pt]{article}

\usepackage[final]{acl}

\usepackage{times}
\usepackage{latexsym}
\usepackage{amsmath,amssymb}

\usepackage[T1]{fontenc}

\usepackage[utf8]{inputenc}

\usepackage{microtype}

\usepackage{inconsolata}

\usepackage{booktabs}
\usepackage{graphicx}
\usepackage{multirow}
\usepackage{multicol}
\usepackage{url}
\newcommand{\textdef}[1]{\textit{#1}}
\newcommand{\word}[1]{``#1''}

\title{On the Distinctive Co-occurrence Characteristics of Antonymy}

\author{Zhihan Cao \\
  Institute of Science Tokyo\\
  School of Computing\\
  \texttt{cao.z.c8a7@m.isct.ac.jp} \\\And
  Hiroaki Yamada \\
  Institute of Science Tokyo\\
  School of Computing\\
  \texttt{yamada@comp.isct.ac.jp} \\\And
  Takenobu Tokunaga \\
  Institute of Science Tokyo\\
  School of Computing\\
  \texttt{take@c.titech.ac.jp} 
  }

\begin{document}
\maketitle
\begin{abstract}
Antonymy has long received particular attention in lexical semantics.
Previous studies have shown that antonym pairs frequently co-occur in text, across genres and parts of speech, more often than would be expected by chance.
However, whether this co-occurrence pattern is distinctive of antonymy remains unclear, due to a lack of comparison with other semantic relations.
This work fills the gap by comparing antonymy with three other relations across parts of speech using robust co-occurrence metrics.
We find that antonymy is distinctive in three respects: antonym pairs co-occur with high strength, in a preferred linear order, and within short spans.
All results are \href{https://github.com/hancules/AntonymySpecificity}{available online}.
\end{abstract}

\section{Introduction}
\label{sec:intro}
Among various semantic relations, antonymy has long received particular attention.
\citet{Charles_1989} proposed \textdef{the co-occurrence hypothesis}, suggesting that adjectival antonymy pairs tend to co-occur within a sentence more often than expected by chance.
Subsequent studies have provided empirical support for this hypothesis~\cite{Justeson_1991}.
This co-occurrence tendency has also been observed in parts of speech~(PoS) other than adjectives~\citep{Fellbaum_1995, Mohammad_2008} and across various genres~\citep{Jones_2005,Jones_2006}.

Prior studies have identified several characteristics of the co-occurrence of antonymy pairs.
One is that co-occurrence is likely to be ordered.
Each antonymy pair often appears in a specific linear order~\citep{Kostic_2015, Kostic_2017, Wu_2022}.
The more frequent antonym tends to precede its less frequent counterpart~\cite{Jones_2002}.
Another characteristic is that antonymy pairs likely co-occur in close proximity.
This proximity typically falls within five-word windows or appears in coordinated constructions~\cite{Jones_2007, Mohammad_2008}.
Therefore, antonymy pairs are characterised by an intra-sentential co-occurrence pattern, which might be distinctive of antonymy.

However, this assumption remains largely untested, as the intra-sentential co-occurrence characteristics of non-antonymy relations have been underexplored.
Some indirect evidence does exist.
For example, hypernymy or holonymy pairs can be retrieved using lexico-syntactic patterns~\citep{Hearst_1992, Girju_2006, Roller_2018}.
Yet these approaches rely on specific constructions rather than reflecting the general co-occurrence patterns within sentences.
As a result, existing work is insufficient to determine whether the co-occurrence pattern uniquely characterises antonymy.

There is also a methodological concern.
Prior studies commonly relied on raw co-occurrence counts or pointwise mutual information~(PMI) of word pairs.
Such metrics are biased toward extremely low or high frequencies that are common due to the sparsity of word frequencies~\citep{Schulte_2008}.
The sparsity also makes many statistical tests~(such as the Chi-square test used by~\citet{Kostic_2017}) unreliable~\citep{Dunning_1993}, raising concerns about whether the co-occurrence observed is truly significant.

If antonymy indeed stands out in its co-occurrence, this distinctiveness needs to be established rather than assumed.
Understanding different semantic relations in natural text is foundational not only to linguistic theory but also to computational modeling. 
Recently, pretrained language models (PLMs) have exhibited a puzzling phenomenon, \textdef{antonymy bias}.
Regardless of architecture and size, PLMs perform substantially better on predicting the antonym of a given word over other relations~\citep{Pitarch_2023, Cao_2024}.
Establishing the distinctive co-occurrence characteristics of antonymy will lay the groundwork for analysing the semantic behavior of PLMs trained on a plain corpus. 

\paragraph{Research Objective}
In summary, although prior studies have suggested that antonymy exhibits distinctive co-occurrence characteristics, existing evidence remains insufficient to validate this assumption.
The present work evaluates and compares the co-occurrence of antonymy and other semantic relations across PoS, using robust and statistically grounded co-occurrence metrics. 
We aim to establish whether antonymy presents a unique pattern of co-occurrence characteristics.

\section{Metrics}
Semantic relations theoretically hold between word lemmas that share the PoS.
We then define metrics at the lemma level. 

\subsection{Strength of Co-occurrence }
The first metric, $G^2$ score~\citep{Dunning_1993}, measures the co-occurrence strength. 
It directly addresses the methodological concerns of prior studies because it is theoretically and empirically robust to data sparsity~\citep{Moore_2004,Evert_2009} and supports a reliable statistical test.

The $G^2$ score of a word pair $(w,v)$ is essentially the logarithm of the likelihood ratio test statistic.
The underlying null hypothesis is that $(w,v)$ co-occurs in the same sentence at a chance level\footnote{The original null hypothesis is that $(w,v)$ co-occurs adjacently by chance. We relax it according to our research objective.}.
$G^2$ scores asymptotically follow a $\chi^2$ distribution with one degree of freedom.
This allows us to interpret the score in two ways:  
1) whether the co-occurrence is statistically significant, and  
2) if it is, a higher score means stronger co-occurrence.

Formally, the $G^2$ score is defined as 
\begin{equation}
    G^2(w,v) = 2 \ \sum_{c\,\in\, C} O_c \log {\frac{O_c}{E_c}},
\end{equation}
where $C$ denotes a set of the possible events regarding the co-occurrence of $(w,v)$ in a sentence: $\{wv, w\bar{v},\bar{w}v, \bar{w}\bar{v}\}$.
Here, $wv$ indicates that both $w$ and $v$ appear in the same sentence; $w\bar{v}$ and $\bar{w}v$ denote that only $w$ or only $v$ appears, respectively; $\bar{w}\bar{v}$ denotes that neither occurs.
$O_c$ and $E_c$ represent the observed and expected number of sentences for each event $c$.
The expected count $E_c$ is calculated under the assumption that $w$ and $v$ occur independently. For example, when $c = wv$, the expected count is computed as $E_c =\frac{|w|\times |v|}{N}$ where $|w|$ and $|v|$ are the number of sentences containing $w$ and $v$, respectively, and $N$ is the total sentence number.

The advantages of $G^2$ scores are three-fold.
First, $G^2$ scores normalise the co-occurrence of two words by their individual frequency and are hence robust to extremely high frequencies.
Second, the use of logarithm mitigates the effect of extremely low frequencies.
Finally, $G^2$ scores incorporate more information than PMI and raw counts by considering both how often a pair co-occurs and how often it does not.
This enables an evaluation of the co-occurrence strength more comprehensively without being biased only by the co-occurred side.

\subsection{Order of Co-occurrence}
\label{sec:order}
Prior studies~\cite[etc.]{Kostic_2015} suggest the order might distinguish antonymy from other relations.
Our second metric evaluates the linear order of a word pair $(w, v)$.
We let $w$ always be the more frequent word and $v$ the less frequent one.
For each sentence where $(w, v)$ co-occurs, we assign an order score of $+1$ if $w$ precedes $v$, and $-1$ otherwise.
Then, for every word pair $(w, v)$, we obtain a list of order scores.
We determine whether there is a preferred order of $(w, v)$ by conducting a binomial test on whether the proportion of $+1$ is significantly greater or less than 0.5.
The order of a pair is defined as the average order score if the pair has a preferred order, and zero otherwise.

\subsection{Distance of Co-occurrence}
Whether the co-occurrence takes place in close proximity may be a distinctive characteristic of antonymy~\cite[etc.]{Mohammad_2008}.
The last metric measures the distance of co-occurrence, defined by the average number of words separating the two co-occurring words in a sentence.

\paragraph{Statistical Tests}
In order to establish whether the antonymy is statistically different from other relations, for each metric, we conduct a Brunner-Munzel test on every possible relation pair.
It generalises the Mann-Whitney U test by relaxing the assumption of equal variances between samples being compared, showing better empirical robustness~\cite{Brunner_2000}.
For statistical tests described above~(test of individual $G^2$ scores, binomial test, and Brunner-Munzel test), the significance level is set at 0.01.

pos   relation
NOUN  HYP-HPO     11664

\section{Data}          
\begin{table}[t!]
\centering\small
\begin{tabular}{l|rrrrr}
\toprule
POS & ANT & HOL & HYP  & SYN  & UNR \\
\midrule
NOUN & 97 &  758    &  11,664 &  1,078  & 7,618 \\
VERB & 74 &     --  &  6,853  &  1,214  & 869\\
ADJ  & 192 &     -- &      -- &   309   & 1,409\\
ADV  &  20 &     -- &      -- &   140   & 104\\
\midrule
Total & 383 & 758 & 18,517 & 2,741 & 10,000 \\
\bottomrule
\end{tabular}
\caption{Counts of lemma pairs.
``--'' means the relation is not defined for the PoS in WordNet.
}
\label{tab:relation_pos_counts}
\end{table}
We evaluate four semantic relations: antonymy (ANT), synonymy (SYN), hyper-hyponymy (HYP), and holo-meronymy (HOL).
The hyper-hyponymy is defined to hold if either word in a pair is a hypernym of the other; similarly, holo-meronymy holds if either is a holonym of the other.
This treatment aligns the definitions of all relations with the symmetric nature of antonymy and synonymy, making all relations comparable.

We retrieve lemma pairs of nouns, verbs, adjectives, and adverbs in any of the four relations from WordNet~\citep{Miller_1995}.
A series of filtering steps is applied in order to ensure that all lemma pairs are of the same linguistic unit level and are lexico-semantically interpretable.

We exclude:
1) pairs containing multi-word expressions, abbreviations, or named entities;  
2) pairs where either word has a frequency of zero or one in WordNet;  
3) pairs where multiple semantic relations exist between the two lemmas;  
4) verb pairs involving linking verbs, auxiliary verbs, or light verbs;  
5) hypernymy pairs with a path length more than two in WordNet hierarchy\footnote{Because hypernymy pairs with a path length of two or less are most intuitive to human judgments~\cite{hyperlex}.
}.
These procedures result in 25,115 lemma pairs.

We estimate the metrics on the Corpus of Contemporary American English~(COCA, \citeauthor{coca},~\citeyear{coca}).
It is designed to reflect the general usage patterns of the English language across balanced genres.

Choosing COCA brings an additional benefit: it is annotated with both PoS tags and lemmas.
The PoS and lemma annotations allow us to efficiently check semantic relations between two co-occurring lemmas using WordNet.

To reduce noise, we exclude sentences shorter than five words, which are often exclamatory or fragmentary.  
This leaves 17,718,403 sentences for analysis.
Among the 25,115 lemma pairs retrieved from WordNet, 22,399 lemma pairs are observed in COCA.

As a control, we additionally randomly sample 10,000 lemma pairs from all lemma pairs that co-occur intra-sententially but are unrelated~(UNR) in any relation defined in WordNet.
The filtering procedures described above are applied to unrelated pairs as well.

We sort the two lemmas in each pair by descending frequency.
Table~\ref{tab:relation_pos_counts} shows statistics of lemma pairs for each relation.
The counts are highly imbalanced across relations, reflecting the sparsity of WordNet~\citep{Cao_2025}.

\section{Results}                
\begin{figure}[t!]
    \centering
    \includegraphics[width=\linewidth]{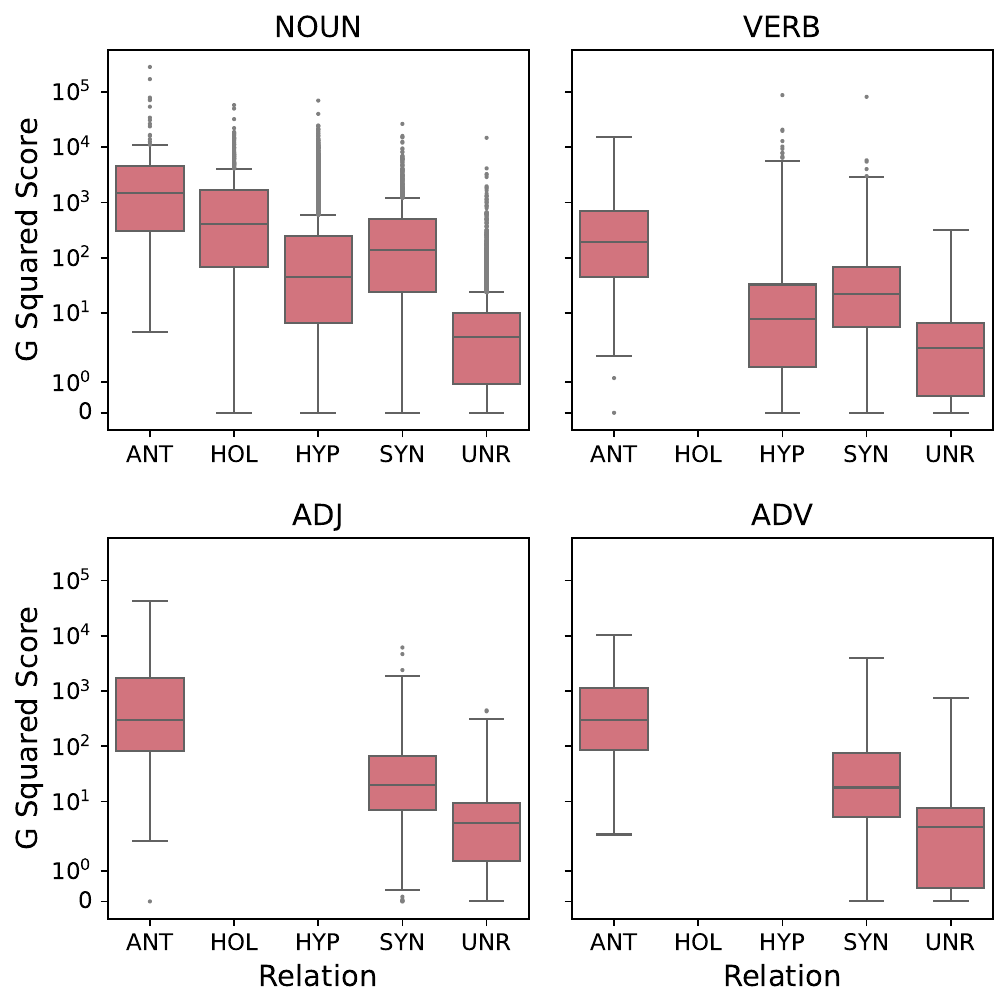}
    \caption{Distribution of $G^2$ scores per PoS and relation with the y axis in logarithmic scale.}
    \label{fig:g2}
\end{figure}

\begin{table}[t!]
\centering\small
\begin{tabular}{l|ccccc}
\toprule
PoS & ANT & HOL & HYP & SYN & UNR \\
\midrule
\multirow{2}{*}{NOUN}
& \textbf{11,144}  & 1,819   & 406  & 628  & 19 \\
& 99\% & 92\% & 75\% & 88\% & 36\%       \\
\midrule
\multirow{2}{*}{VERB} 
& \textbf{915}   & --  & 83   & 164  & 7  \\
& 91\% & --  & 52\% & 72\% & 24\%       \\
\midrule
\multirow{2}{*}{ADJ}  
& \textbf{2,309}  & --  & -- & 120  & 9  \\
& 97\% & -- & -- & 76\% & 36\%       \\
\midrule
\multirow{2}{*}{ADV}  
& \textbf{1,471}  & -- & -- & 128   & 14    \\
& 95\% & -- & -- & 71\%  & 33\%       \\
\bottomrule
\end{tabular}
\caption{Average $G^2$ scores (above) and percentages of significant cases (below).}
\label{tab:g2_combined}
\end{table}
Table~\ref{tab:g2_combined} presents the average $G^2$ scores and the percentage of pairs with significant scores.
In this table and all subsequent tables, boldface indicates relations whose score differs significantly from all other relations within the same PoS for the corresponding metric.
Antonymy pairs consistently yield both the highest $G^2$ scores~(ranging from 915 to 11,144) and the largest percentage~(at least 91\%) of significant co-occurring pairs across all PoS.
Figure~\ref{fig:g2} shows the distribution of $G^2$ scores per relation and PoS.
The first and third quantiles of $G^2$ scores for Antonymy are higher than those of all other relations across PoS.
These results confirm that antonyms not only co-occur more frequently than expected by chance, but also with greater strength than other relations.

We acknowledge that antonymy tends to have extremely high $G^2$ scores.
For nominal antonymy, there are even two pairs, (\word{child},\word{parent}) and (\word{man},\word{woman}), whose $G^2$ scores are greater than 100,000.
Such extremely high outliers might have inflated the average $G^2$ scores to 11,144 for nominal antonymy, making it the highest among all PoS and relations.

\begin{table}[t!]
\centering\small
\begin{tabular}{l|rrrrr}
\toprule
PoS & ANT & HOL & HYP & SYN & UNR \\
\midrule
\multirow{2}{*}{NOUN}
&\textbf{.13} & .05 & .02 & .02  & .01  \\
& 97\% & 96\% & 93\% & 90\% & 33\%  \\
\midrule
\multirow{2}{*}{VERB}
& \textbf{.11} & --   & .03 & .02  & .02  \\
& 94\% & -- & 88\% & 81\% & 40\%  \\
\midrule
\multirow{2}{*}{ADJ}
&\textbf{.11} & --   & --    & .04  & .00 \\
& 91\% & -- & -- & 61\% & 21\%  \\
\midrule
\multirow{2}{*}{ADV}
& \textbf{.29} & --   & --    & .05  & .03  \\
& 95\% & -- & -- & 87\% & 29\%  \\
\bottomrule
\end{tabular}
\caption{Average order scores~(above) and the percentages of pairs with a preferred order among all significantly co-occurring pairs~(below).
}
\label{tab:order}
\end{table}
Table~\ref{tab:order} presents the results of the order preference.
Antonymy pairs show a strong tendency to prefer a specific order.
Across all PoS, more than 90\% of the co-occurring antonymy pairs have a preferred order, typically with an average order score above 0.10.
The average order scores for antonymy are overall significantly larger than other relations across PoS.
This indicates that the more frequent antonym slightly tends to precede the less frequent counterpart.
For other relations, a preferred order exists at the pair level but is not consistent across pairs, resulting in no clear pattern at the relation level.

\begin{figure}[th!]
    \centering
    \includegraphics[width=\linewidth]{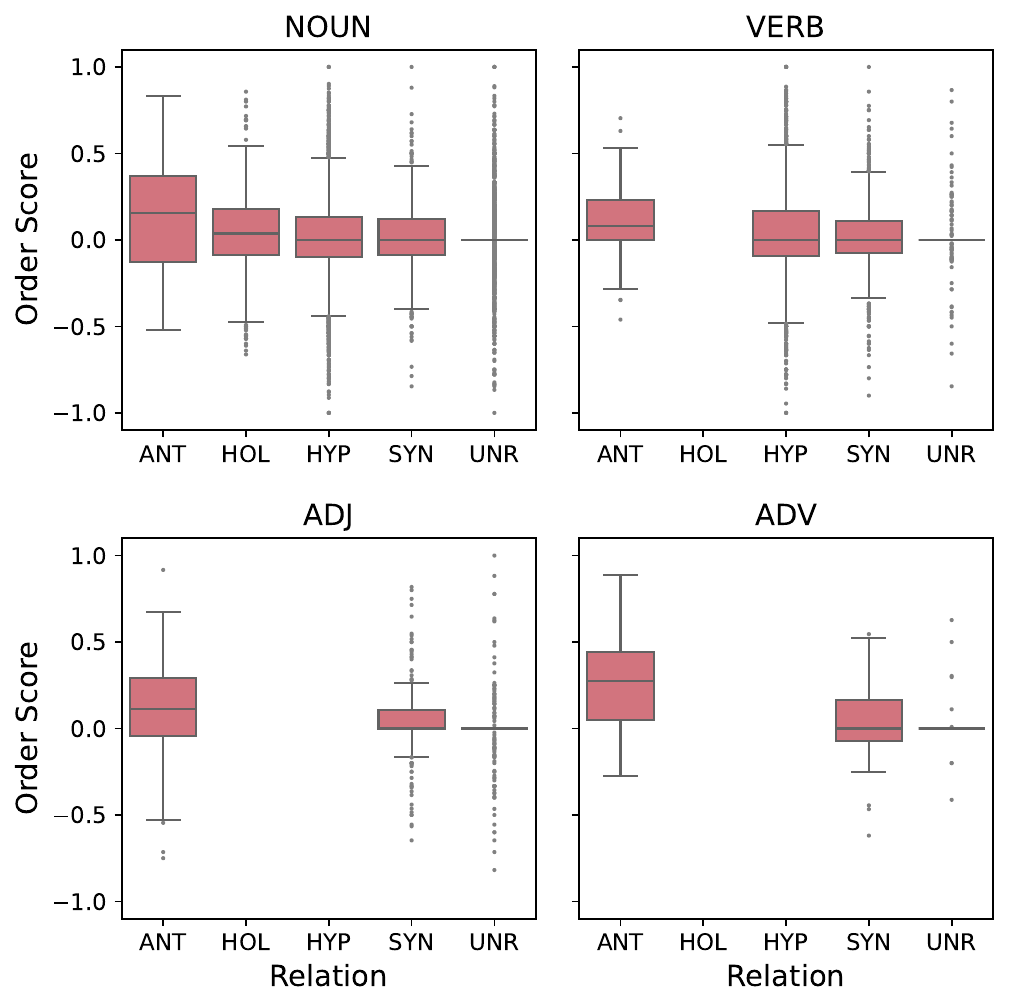}
    \caption{Distribution of order scores per PoS and relation.}
    \label{fig:order}
\end{figure}
Figure~\ref{fig:order} shows the distribution of order scores per relation and PoS.
For non-antonymy relations, the range of order scores tends to be around 0, confirming that there is no clear co-occurrence order preference\footnote{
Note that this finding is based on our definition of relations, in which holonymy and meronymy are not distinguished, nor are hypernymy and hyponymy. If these distinctions are made, holonyms tend to precede meronyms to a moderate degree. Detailed results are provided in Appendix~\ref{sec:ordered}.
}.

Table~\ref{tab:avg_distance} shows the co-occurence distance.
\begin{table}[t!]
\centering\small
\begin{tabular}{l|ccccc}
\toprule
PoS & ANT & HOL & HYP & SYN & UNR \\
\midrule
NOUN & \textbf{15} & 22 & 26 & 25 & 30 \\
\midrule
VERB & \textbf{18} & --   & 28 & 25 & 50 \\
\midrule
ADJ  & \textbf{14} & --   & --   & 17 & 25 \\
\midrule
ADV  & \textbf{11} & --   & --   & 20 & 27 \\
\bottomrule
\end{tabular}
\caption{Average distances between significantly co-occurring pairs.
}
\label{tab:avg_distance}
\end{table}
Antonyms co-occur more closely than other relations.
On average, antonyms co-occur within 18 words, which is shorter than the span of pairs in other relations across PoS.
\begin{figure}[th!]
    \centering
    \includegraphics[width=\linewidth]{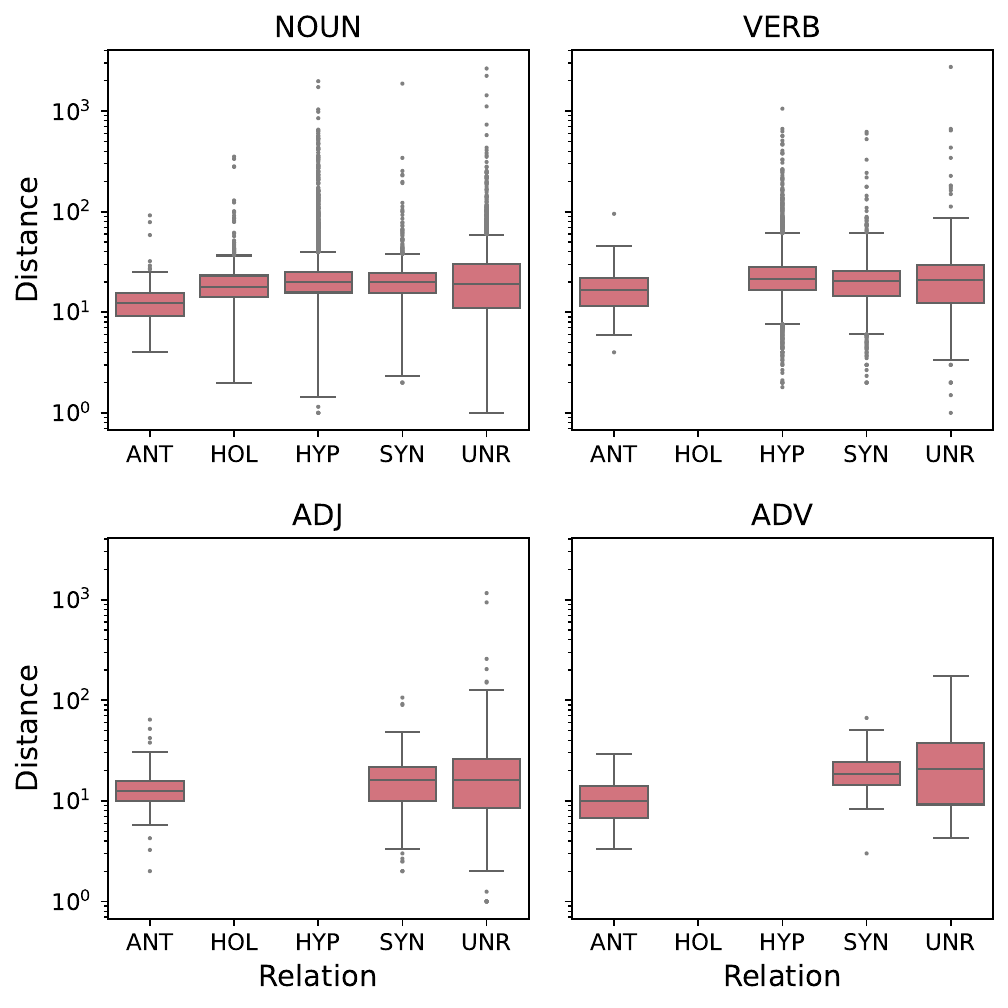}
    \caption{Distribution of distances per PoS and relation with the y axis in logarithmic scale. }
    \label{fig:distance}
\end{figure}
Figure~\ref{fig:distance} shows the distribution of distances per relation and PoS.
For antonymy, the third quantile and the maximum of co-occurrence distance are consistently the lowest among all relations across PoS, confirming that antonymous pairs tend to occur close in text.

In short, compared to other relations, antonymy pairs co-occur
1) with greater strength, 
2) in the order where the more frequent word tends to precede, and 
3) at a shorter distance.
Our findings align with previous observations on antonyms, and extend them to other relations and PoS, revealing the distinctive co-occurrence characteristics of antonymy.

\section{Discussion}
\paragraph{Cross-PoS Analysis}
The strong co-occurrence of antonymy pairs is often attributed to semantic contrast.
Antonymy pairs denote two opposite ends of a scale or a dichotomy in a situation~\cite{Cruse_1986}, which might help highlight semantic contrast and hence lead to their significant co-occurrence~\cite{Jones_2006}.
If the significant co-occurrence is indeed associated with semantic contrast, it should persist after derivation, which usually changes the PoS but not the core meaning.
To test this, we examine whether co-occurrence significance is retained after derivation.
Among the total of 192 adjectival antonymy pairs, we identify 18 pairs from which 19 non-adjective antonymy pairs are derived~(e.g. (\word{strong},\word{weak}) to (\word{strongly},\word{weakly}))\footnote{
This is done using the information in WordNet. Details can be found in Appendix~\ref{sec:der}.
}.
The 18 original adjectival antonymy pairs co-occur significantly; 18 out of the 19 derived non-adjectival antonymy pairs also exhibit significant co-occurrence.
For other PoS and relations, we see a similar pattern.  
Among all non-adjectival or non-antonymy pairs that significantly co-occur, 85\% of their derived pairs sustain the significant co-occurrence.
These findings suggest that significant co-occurrence is robust under derivational change, aligning with the idea that co-occurrence might stem from the semantic contrast.

\paragraph{Cross-relation Analysis}
Antonymy is considered to differ from hypernymy, holonymy, and synonymy in nature because it is not just semantic but also lexically constrained~\cite{Justeson_1991}.
For example, \word{hot} is in antonymy with \word{cold} but not with \word{chill}.
We verify to what extent each relation is lexically constrained.
For each relation, we calculate how many lemmas the more frequent lemma is associated with on average~\footnote{
The Appendix~\ref{sec:app} presents the details.
}.
For antonymy, the frequent lemma is associated with only one lemma on average, whereas for other relations, it ranges from 1.7 to 6.7.
This result suggests that antonymy involves the strongest lexical pairing, which might enhance their suitability for parallel constructions and, in turn, contribute to their frequent co-occurrence in texts.

\paragraph{Relating Language Models}
Pretrained language models perform substantially better on antonymy across model architectures and sizes~\citep{Pitarch_2023, Cao_2024}.   
The distinctive co-occurrence characteristics of antonymy may provide signals that facilitate learning antonym pairs during pretraining.
However, as prior studies commonly use sentence completion tasks in evaluation, models might take advantage of those intra-sentential co-occurrence characteristics, using them as a shortcut~\cite{Du_2023} rather than relying on knowledge of antonymy.
Given this, our findings highlight the need to disentangle the extent to which PLMs rely on such distributional clues from the extent to which they generalise beyond them.

\section{Conclusion}
This work presents an initial exploration of the intra-sentential co-occurrence of antonymy in comparison to other semantic relations across PoS.
We find that antonymy is consistently distinctive from all other relations across all metrics used.
Our findings establish a robust empirical foundation for the distributional nature of antonymy, offering a solid empirical basis for its future analysis.

\section*{Limitations}
We only characterise the co-occurrence of semantically related words in a quantitative manner and do not address the qualitative evaluation.
We hence can not answer in which lexical and dependency constructions two words are likely to occur.
Furthermore, as an exploratory study, we did not apply word sense disambiguation to a sentence but only used the lemma and PoS of each word as a shallow filtering.
We chose not to apply word sense disambiguation in alignment with the previous studies which are done on lemma level.
We only studied COCA, and hence, future work should focus on a specific pretraining corpus and examine how PLMs trained on it become biased toward antonymy.

\section*{Acknowledgments}
This work was supported by JSPS KAKENHI Grant Number JP25KJ1271. 
We would like to extend our sincere gratitude to Prof. Simone Teufel for valuable discussions.  
We would also like to express our appreciation to Wanjing Cao, Huaqing Qi, and Shuqin Zhang for their support throughout this work.

\bibliography{starsem}

\appendix
\section*{Appendix}
\section{Ordered Scores for Asymmetric Relations}
\label{sec:ordered}
\begin{figure}[th!]
    \centering
    \includegraphics[width=\linewidth]{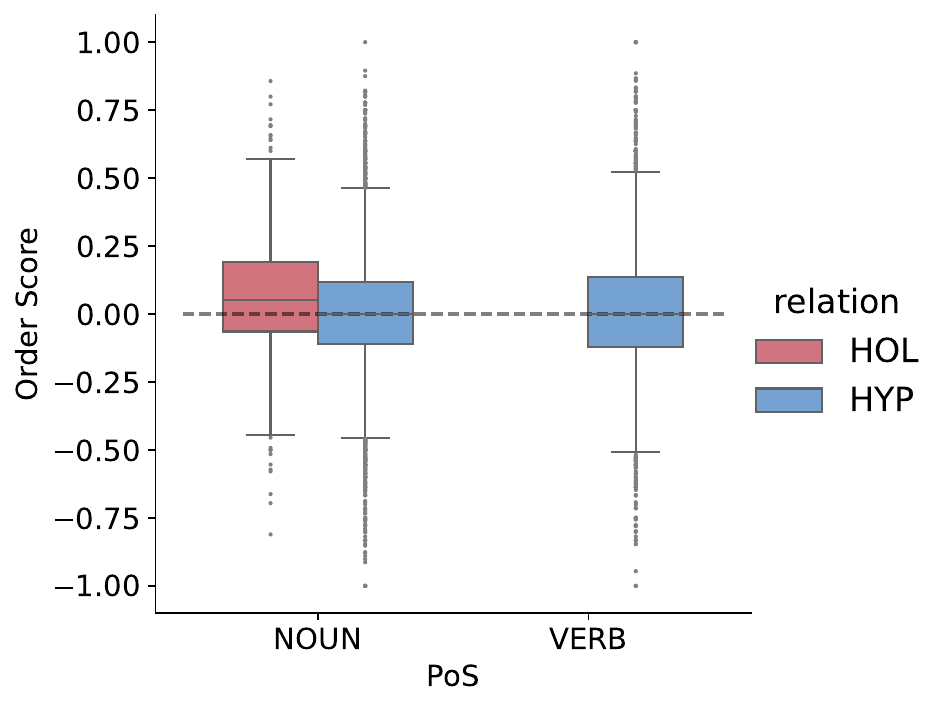}
    \caption{Distribution of order scores of asymmetry relations per PoS. }
    \label{fig:ordered_order}
\end{figure}

Figure~\ref{fig:ordered_order} presents the order scores of asymmetric relations.
We redefine the order score for asymmetric relations as follows.
A score approaching 1 indicates that a holonym or hypernym precedes its meronym or hyponym, respectively, while a score close to -1 indicates the reverse.
For meronymy, more than 50\% of the order scores are positive.
The results show that holonyms generally precede their meronyms.
In contrast, for hypernymy and hyponymy, the order score distribution centers around 0, indicating that the ordering tendency is less clear in the hypernymy/hyponymy relations.

\section{Derived Pairs Statistics}
\label{sec:der}
\begin{table}[ht!]
    \centering\small
    \begin{tabular}{ccccr}
    \toprule
    \multicolumn{2}{c}{PoS} & \multicolumn{2}{c}{Rel.} & \multirow{2}{*}{Count}\\
    Orig. & Derv. & Orig.  & Derv.  &  \\
    \midrule
    ADJ  & ADV  & ANT      & ANT      &  13~~(~12) \\
    ADJ  & ADV  & SYN      & SYN      &  8~~(~~~7)  \\
    ADJ  & NOUN & ANT      & ANT      &  5~~(~~~5)  \\
    ADJ  & NOUN & SYN      & SYN      &  1~~(~~~1)  \\
    ADJ  & NOUN & SYN      & HYP      &  8~~(~~~7)  \\
    ADJ  & VERB & ANT      & ANT      &  1~~(~~~1)  \\
    ADJ  & VERB & SYN      & SYN      &  1~~(~~~1)  \\
    NOUN & VERB & ANT      & ANT      &  2~~(~~~2)  \\
    NOUN & VERB & SYN      & SYN      &  15~~(~12) \\
    NOUN & VERB & HYP      & HYP      & 111~~(~90) \\
    NOUN & VERB & SYN      & HYP       & 46~~(~44) \\
    \midrule
    \multicolumn{4}{c}{TOTAL} & 211~(182) \\ 
    \bottomrule
    \end{tabular}
    \caption{Counts of original~(Orig.) and derived~(Derv.) POS pairs and their corresponding relations. The number of pairs that sustain significance after deriviation are in parenthesis.}
    \label{tab:derivation_analysis}
\end{table}
In WordNet, some lemmas are linked to their derived lemmas.
For each lemma pair $(w,v)$ that is semantically related and significantly co-occurs, we retrieve all derived lemmas $w_d$ of $w$ and $v_d$ of $v$ using the information in WordNet.
For every $(w_d,v_d)$, we check whether it forms a semantic relation and appears in our data.
If so, we then examine whether it sustains the co-occurrence significance.

Table~\ref{tab:derivation_analysis} presents the results.
The semantic usually persists after derivation, particularly for antonymy and hyper-hyponymy.
Synonymy pairs sometimes become hyper-hyponymy, reflecting the established similarity between synonymy and hypernymy~\citep{hyperlex, Cao_2025}.

\section{Associated Lemma Counts}
\label{sec:app}
\begin{table}[t!]
\centering\small
\begin{tabular}{c|ccccc}
\toprule
POS & ANT & HOL & HYP & SYN & UNR\\
\midrule
NOUN & 1.0 & 1.9 &  5.8 &  1.5 & 2.3 \\
VERB & 1.1 & --  &  6.9 &  1.9 & 1.5 \\
ADJ  & 1.0 & --  &  --  &  1.4 & 1.6 \\
ADV  & 1.0 & --  &  --  &  1.5 & 1.2\\
\midrule
Micro AVG
& 1.0 & 1.9 & 6.7 & 1.7 & 2.2 \\
\bottomrule
\end{tabular}
\caption{The average number of lemmas that are associated with the more frequent lemmas per relation and PoS.}
\label{tab:share}
\end{table}
For each relation, we count how many lemmas are associated with the more frequent lemma on average.
In other words, for all pairs $(w,v)$, we retrieve pairs that share $w$ and compute the average number of associated $v$s per $w$.
Table~\ref{tab:share} presents the results.

Antonymy pairs usually have a unique associated antonym for each more frequent lemma $w$, as the average number of associated lemmas ranges only from 1.0 to 1.1.
In holo-meronymy and synonymy, the frequent lemma is associated with slightly more lemmas than in antonymy, with the averages ranging from 1.4 to 2.0.
In hyper-hyponymy, the frequent lemma is typically associated with around 6 lemmas.
Hence, antonymy is the most lexically constrained among the other relations studied here.

\end{document}